# Formalizing and Guaranteeing* Human-Robot Interaction

Hadas Kress-Gazit[1], Kerstin Eder, Guy Hoffman, Henny Admoni, Brenna Argall, Ruediger Ehlers, Christoffer Heckman, Nils Jansen, Ross Knepper, Jan Křetínský, Shelly Levy-Tzedek, Jamy Li, Todd Murphey, Laurel Riek, Dorsa Sadigh

Robot capabilities are maturing across domains, from self-driving cars, to bipeds and drones. As a result, robots will soon no longer be confined to safety-controlled industrial settings; instead, they will directly interact with the general public. The growing field of Human-Robot Interaction (HRI) studies various aspects of this scenario – from social norms to colaborative manipulation to human-robot teaming and more. Researchers in HRI have made great strides in developing models, methods, and algorithms for robots acting with and around humans [1], but these "computational HRI" models and algorithms generally do not come with formal guarantees and constraints on their operation. To enable human-interactive robots to move from the lab to real-world deployments, we must address this gap.

Demonstrating trustworthiness in various forms of automation through formal guarantees has been the focus of efforts in validation, verification and synthesis for years. For instance, control systems on board of an aircraft have to meet guarantees, such as correctly handling transitions between discrete modes of operation (e.g., take-off, cruise, landing), simultaneously providing a guarantee on *safety* (e.g., not being in both take-off and landing modes at the same time) and *liveness* - the ability to eventually achieve a desirable state (e.g., reaching cruise altitude). "Formal methods", such as model checking and theorem proving, play a central role in helping us understand when we can rely on automation to do what we have asked of it; they can be used to create correct-by-construction systems, provide proofs that properties hold, or find counterexamples that show when automation might fail.

Formalizing HRI can enable the creation of trustworthy systems and, just as importantly, support explicit reasoning about the context of guarantees. First, writing formal models of aspects of HRI would enable verification, validation, and synthesis, thus providing some guarantees on the interaction. Second, it is unrealistic to verify a complete human-robot system due to the inherent uncertainty in physical systems, the unique characteristics and behaviors of people and the interaction between systems and people. Thus, a formal model requires us to articulate explicit assumptions regarding the system, including the human, the robot, and the environments in which they are operating. Doing so exposes the limits of the provided guarantees and helps in designing systems that degrade gracefully when assumptions are violated.

In this article we discuss approaches for creating trustworthy systems and identify their potential use in different HRI domains. We conclude with a set of research challenges for the community.

---

[1] Corresponding author, Cornell University hadaskg@cornell.edu

## Techniques for demonstrably trustworthy systems

We divide the techniques for gaining confidence in the correctness of a system into four approaches: synthesis of correct-by-construction systems, formal verification at design time, runtime verification or monitoring, and test-based methods. Common to all of these approaches is the need to articulate *specifications*, descriptions of what the system should and should not do. Specifications typically include both safety and liveness properties and are defined in a formal language, for example temporal logics over discrete and/or continuous states, or satisfiability modulo theories (SMT) formulas (e.g. [2]).

The four approaches outlined below are listed in decreasing order of exhaustiveness and, as a result, of computational complexity. The less exhaustive approaches can typically handle more complex systems at a greater level of realism. Synthesis is the most computationally expensive approach and requires the coarsest abstraction but can automatically create a system with guarantees, whereas test-based methods can handle the most complex systems but do not provide formal guarantees regarding the satisfaction of the specifications. In practice, a combination of techniques is required as no single technique can be relied upon on its own [3].

**Synthesis** is the process of automatically generating a system from the specifications. In the context of robotics, there are different techniques for doing so [4] – from offline generation of state machines or policies satisfying discrete and probabilistic temporal logic specifications, through online receding horizon optimization for continuous temporal logics, to online optimization with SMT solvers.

**Formal verification** techniques span methods that exhaustively explore the system (model checking, reachability analysis [2]) to those that reason about the system using axioms and proof systems (theorem proving [5]). Techniques vary from deterministic, worst case analysis, to probabilistic reasoning and guarantees, and from discrete spaces to continuous ones. Such methods are typically applied at design time and either determine that the specification is met in every possible system behavior, or provide a counterexample – a system execution that violates the specification – which may then be used to further refine the design or the specification.

**Runtime monitoring** is the process of continuously checking the correctness of the system during execution using *monitors* that check specifications, either created automatically through synthesis or manually [6]. This type of verification is, in a sense, the most lightweight way of integrating formal methods into a design. It does not alter the design, but enables the detection of failures or the deviation from expected/formalized behavior, to allow shutting down the robot or switching into a safe mode. An additional benefit of runtime-checkable specifications is that they allow us to "probe" the system at design time using, e.g., statistical model checking [7].

**Test-based** methods complement formal methods during verification and validation. In particular, simulation-based testing [8] can expose the system under test to stimuli that are more realistic than the often highly abstracted scenarios that can be verified formally. From a performance point of view, simulation-based testing can reach verification goals faster and with less effort than conventional testing in the real world. *Coverage* is a measurement of verification progress, allowing engineers to track the variety of tests used during testing and how effective they are in achieving verification goals. Assertion monitors act as test oracles, much like the monitors used for runtime verification. Model-based testing is a specific technique that asserts the conformance of a system under test to a given formal model of that system [9]. This is particularly important when guarantees or code generation rely on the correctness of a model.

Validation, verification and synthesis techniques are always related to a given specification. These specifications can never cover the full behavior of a physical system in the world; rather, they include assumptions and abstractions to make the problem tractable. Therefore, guarantees are provided with respect to the specification, enabling us to gain confidence in the overall correctness of the system, and allowing us to narrow down the sources of problems as well as to understand the constraints that limit deployment.

**HRI domains and their unique challenges**
Many HRI domains could benefit from formal methods, and each domain brings about unique challenges:

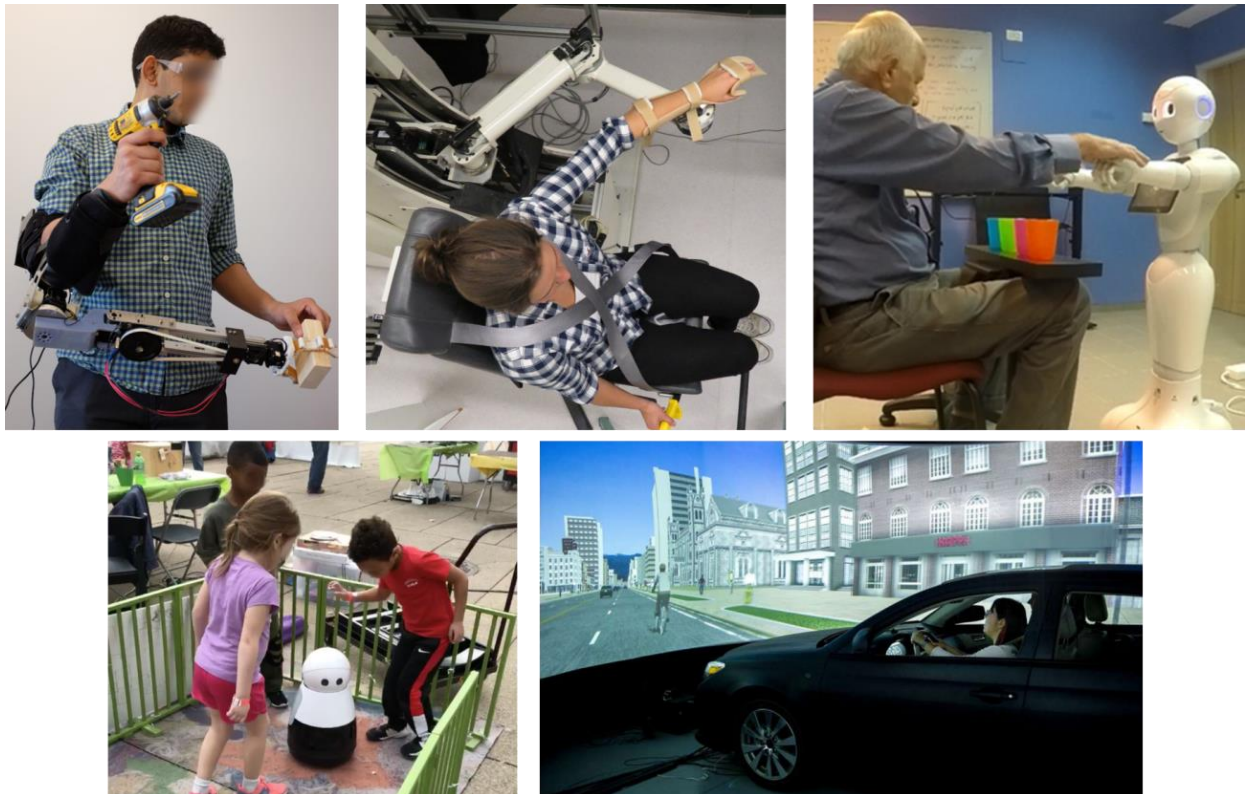

**Figure 1: Domains of HRI that could benefit from formal methods**. Clockwise from top left, physical HRI (construction), physical HRI in the healthcare domain (rehabilitation), cognitive healthcare, autonomous driving, and social HRI.

**Physical HRI** involves systems in which the physical states of an automation interact with physical states of a human [10], for example, a robotic wheelchair carrying a person, or a construction assistant robot carrying a heavy load together with a person. In addition to physical states interacting, their *internal* states interact, since both the robot and the human often have a model of the task they are working to achieve, and a model of each other. For example, in a setting where rehabilitation robots assist an individual with motion, the robot may be responsible for physical safety (e.g., keeping someone upright) while simultaneously maximizing therapy benefit, requiring it to stay out of the way as much as possible. Thus, the system is tasked to assist, but not to overassist. This fundamental tension between the two purposes of the automation with respect to the human leads to challenging questions in terms of specification (e.g., how does one articulate the notion of safety while avoiding overassisting) and verification (e.g., how to prove that the control methods satisfy the specification).

**Healthcare Robotics**: There are a variety of robots being developed to assist people with activities of daily living, including physical mobility, manipulation, medicine reminders, and cognitive support. Robots are also developed to support clinicians, caregivers, and other stakeholders in healthcare contexts [11]. For example, physically assistive robots, such as exoskeletons and robotic prostheses, can help individuals perform movements, such as walking and reaching, and socially assistive robots can help individuals engage in positive health behaviors, such as exercise and wellness [12]. People have different abilities and disabilities that may change over short and long time horizons. Therefore, modelling a person's ability and personalizing the system is crucial for creating successful HRI systems in the healthcare domain.

**Autonomous Driving**: Recent years have seen significant advances in autonomous driving. As these fully or semi-autonomous vehicles appear on the road, challenges arise due to interactions with humans. Humans, in the context of autonomous driving, fall into three main categories: 1) human drivers or riders in the autonomous vehicle; 2) human drivers of other vehicles around the autonomous car; and 3) pedestrians or bicyclists interacting with autonomous vehicles on roads. An obvious specification in this domain is safety - no collisions. However, that specification is not enough; when merging onto a highway, the safest course of action is to wait until there are no other vehicles close by. On busy roads this is not a reasonable course of action. Therefore, the specification needs to go beyond addressing the challenges of driving a single vehicle, and formalize desirable behavior when cars interact with other vehicles and road users [13]. The challenges of this domain are to model and validate acceptable road behavior, reason about expected and unexpected behaviors by people in all the above categories, and provide certification, diagnosis, and repair techniques that will enable autonomous vehicles to drive on our roads.

**Social Collaboration**: In addition to the contexts listed above, there are many instances in which humans and robots will engage in predominantly social, rather than physical, interactions [14]. For example, an information kiosk robot at an airport might engage in a conversation to get a person to where they want to go. Social collaborations across many domains can be characterized by the use of social actions, such as verbal and nonverbal communication, to achieve a shared goal. Social collaboration typically requires the agents involved to maintain a *theory of mind* about their partners, identifying what each agent believes, desires, and aims to achieve. In social collaboration, it is important that the robot follows social norms and effective collaboration practices, for example not interrupting the speaker and providing only true and relevant information [15]. If a robot fails to follow such conventions, it risks failing at the collaboration due to lack of trust or other social effects. One major challenge of formalizing social collaborations is how to encode social norms and other behavior limitations as formal constraints. Researchers interested in verification or synthesis of social collaborations will have to identify which social behaviors and which elements of the task are important for the collaboration to succeed.

**Work in formalizing HRI**
Researchers in computational HRI [1] have developed models for human behavior, for human-robot collaboration and interaction, and algorithms that have been demonstrated in various HRI domains. Whereas these approaches are evaluated qualitatively and quantitatively, the HRI research community has not often formalized what constitutes "correct" behavior. Generally speaking, there are very few examples of formal specifications, or algorithms that can verify or synthesize such specifications.

In the past few years, collaborations between HRI researchers and researchers studying formal methods, verification, and validation have begun to address the challenge of formalizing specifications and creating demonstrably trustworthy HRI systems. Some efforts have explored Linear Temporal Logic as a formalism to capture and verify norms in an interaction [16] and to synthesize human-in-the-loop cyber-physical systems [17]. Other examples include using Satisfiability Modulo Theories for encoding social navigation specifications [18], Signal Temporal Logic for handover behavior [19], and automata-based assertion monitors for robot-to-human handover tasks [20].

Other researchers have focused on socio-cyber-physical systems for instance by including human factors into cyber-physical systems, ranging from specific roles of humans, their intentions, legal issues, and level of expertise [21]. Other work models an assisted-living scenario as a Markov decision process [22] making use of the probabilistic model checker PRISM [23].

**Challenges for the research community**
Work described above suggests the promise of introducing formal methods techniques into HRI domains. That said, creating and reasoning about trustworthy HRI requires addressing the unique aspects of HRI and rethinking current approaches to verification, validation, and synthesis of systems. In this section, we distill three unique aspects of HRI research posing a challenge for formal methods: designing useful specifications for HRI, dealing with the expected adaptation of the human to the automated system, and handling the inherent variability of humans. For each challenge domain, we identify high-priority research directions that could drive progress toward creating trustworthy HRI systems.

**Designing Formal Specifications for HRI:** Whenever verifying, testing or synthesizing a system, one needs to formalize the system by defining the state space of the model and the specification of interest. For example, in the context of autonomous cars obeying the law and social conventions, the state space may include the position and velocity of the car and any other cars in the environment. The specification may represent a requirement of the form "the car never exceeds the speed limit and always maintains a safe distance from all other cars". In the context of HRI, designing useful specifications raises several research questions:

- **What should be the space of specifications?** In HRI, simply modelling the physical state of the robot and the human is usually not enough. The physical state does not capture requirements such as avoiding over-assisting a person or maintaining social and cultural norms. We need to create richer spaces that enable writing such specifications while balancing the complexity of the algorithms that will be used for verification and synthesis in these spaces.
- **How to write specifications that capture trust?** A human will only trust a robot to react in a safe way if it obviously and demonstrably does so. Hence, the robot needs to not only be safe but also be *perceived* as safe, which may require a considerable safety margin. On the other hand, when the interaction involves shared human-robot control, equally important to the idea of humans trusting the robot is the notion of whether and to what extent the robot can trust the human. This plays a role in determining under what circumstances the robot should step in, and in what manner. Particularly in safety-critical scenarios, and when the robot is filling a gap in the human's own capabilities, reasoning about trust in the human is key. Critical factors are to measurably assess (1) the human's ability to actually perform the task, and (2) the current state of the human, for instance accounting for levels of fatigue. These notions of trust go beyond typical safety and liveness specifications and require specification formalisms that can capture them.
- **What should be the definition of failure?** Beyond failure with respect to physical safety that is well studied in the literature, interaction failures may have varying impacts. A small social gaffe such as intruding on personal space may not be an issue, but a large mistake like dropping a jointly manipulated object might have a long-term effect on interaction. We need to be able to define specifications that capture the notion of social failure and develop metrics or partial orders on such failures, so that the systems can fail gracefully.
- **How to formalize the human's behavior during an interaction?** A common technique in verification is assume-guarantee reasoning, where a system's behavior is verified only under the assumption that its input satisfies a well-defined specification. If the input violates the assumption, the system behavior is no longer guaranteed. Given our understanding and observations of human-human and human-robot interaction, a challenge for synthesizing and verifying HRI is to formalize the assumptions on the behavior of the human, who provides the input of the HRI system, in a way that supports verification, is computationally tractable, and captures the unique characteristics of humans.

**Adapting to Human Adaptation**: During interaction, humans and robots will engage in mutual adaptation processes [24]. For example, people become less cautious operators of machines (e.g., cut corners, give narrower berth to obstacles) as they become more familiar with them. Therefore, any models used to represent the interaction and reason about it must capture this adaptation. To complicate matters, the temporal adaptation may occur at different time scales: short time scales, for example morning vs. evening fatigue, and longer time scales, for example improvement or deterioration in functional ability over months [24], [25]. Changing models in itself makes formalizing HRI more complicated, but it is the diversity of the ways humans adapt to a task and a team mate that makes their accurate modeling even more challenging. This property brings up the following research challenges:

- **Which mathematical structures can capture non-stationary models?** Mutual adaptations are common in human-human interaction. For example, humans build conventions when communicating with each other through repeated interactions using language or sketches [26] – studying these interactions and formalizing them can form the basis for new HRI models. When developing such models, an important consideration is how to capture the different time scales of adaptation.
- **How can the robot detect and reason about the human's adaptation?** As the human adapts to the interaction, their behavior (and thus the input to the interaction) may change. For example, people may become less emotionally expressive as the novelty of the interaction wears off or they may give less control input as they trust the autonomy of the system more; this in turn creates a challenge at runtime when a robot is attempting to ascertain how the human adapted. We need to develop runtime verification algorithms that can detect such adaptation and influence the interaction.
- **How to model feedback loops?** As the robot and the human adapt to each other, it is important to reason about the positive and negative feedback loops that emerge and their effect on the resulting interaction. These feedback loops can take the human-robot systems to desirable or undesirable equilibria. For example, the difference between driving cultures around the world may be explained by repeated interactions between drivers causing behavioral feedback loops, leading to emergent locally distinct conventions. We need to study the long term behavior of repeated interactions and adaptations, and verify the safety of the resulting emergent behaviors.

**Variability among Human Interactants**: While we can reasonably assume that the model of a particular type of robot is the same for all robots of that type, there does not exist a model of a "typical" human – one size does not fit all. Even identifying the proper parameters or family of parameters that encapsulate the types of variability in people is a seemingly impossible task. People differ across backgrounds, ages and abilities, which raises the important question of how much to personalize the model and specification to a specific individual or population:

- **Can we identify general specifications for which one simple human model is enough?** Is it possible to create a basic, human-centric and application agnostic model of human behavior that indicates a basic specification such as loss of engagement of a human in the interaction? Such a generic model can detect behavior outside the expected, for example distraction or lack of attention, and could be used to trigger safety measures irrespective of the specific application area. A current example for such a model is used in driver assist systems; they measure where the driver is looking, suggesting the driver take a break if they detect staring or lack of eye movement - universal signs for sleepiness.
- **What levels of personalization are needed?** Refining the research question above, it is important to study not only the formalisms that allow models and specifications to be personalized but also to what extent personalization is required for smooth interaction, what are the trade-offs between complexity of the model and improved interaction, and what are the metrics that enable reasoning about the trade-offs. For this purpose, models of mental representations (e.g. levels of cognitive control for error-free decision making [27]) could be useful.
- **How to model the human's ability level?** The interaction should be appropriate for the ability level of the person; when the human is better off completing a task on their own, too much assistance is not desirable, for example in therapeutic and educational settings. In other cases , too little assistance can be frustrating and lead to disengagement. It is important to model both the ability and the modes of interactions that are most appropriate for each task.
- **How to formalize experiential considerations?** People from different backgrounds may have different assumptions (e.g., [28]) and expectations (e.g., [29]) from robots, and may perceive the interaction with the robot differently. Since meeting user expectations is important for fostering trust between the human and the robot [30], [31], the personalization of the interaction should consider the experiential background of the user, who may expect the robot to be, for example, more assertive and active, or more meek and passive.

## **Conclusion**

As robots begin to interact closely with humans, we need to build systems worthy of trust regarding both the safety and the quality of the interaction. To do so, we have to be able to formalize what a "good" interaction is, and we need algorithms that can check that a given system produces good interactions or even synthesize such systems. To make progress, we must first acknowledge that a human is not another dynamic physical element in the environment, but has beliefs, goals, social norms, desires, and preferences. To address these complexities, we must develop models, specifications, and algorithms that make use of our knowledge about human behavior to create demonstrably trustworthy systems. In this paper, we identified a number of promising research directions and we encourage the HRI and formal methods communities to create strong collaborations to tackle these and other questions towards the goal of trustworthy HRI.

**Acknowledgment**: This paper is a result of fruitful discussions at the Dagstuhl seminar on Verification and Synthesis of Human-Robot Interaction [32]. The authors thank all fellow participants and the Schloss Dagstuhl – Leibniz Center for Informatics, for their support.